\pdfoutput=1

\documentclass[10pt,twocolumn,letterpaper]{article}

\usepackage{cvpr}
\usepackage{times}
\usepackage{epsfig}
\usepackage{graphicx}
\usepackage{amsmath}
\usepackage{amssymb}
\usepackage{booktabs}
\usepackage{paralist}
\usepackage{textcase}
\usepackage{float}
\usepackage{stackengine}
\usepackage{subcaption}
\usepackage[pagebackref=true,breaklinks=false,colorlinks=true,bookmarks=false]{hyperref}
\usepackage[super]{nth}
\usepackage[utf8]{inputenc}
\cvprfinalcopy

\def\cvprPaperID{10212} 

\begin{document}


\newcommand{\figref}[1]{Figure~\ref{#1}}
\newcommand{\secref}[1]{Section~\ref{#1}}
\newcommand{\tabref}[1]{Table~\ref{#1}}

\newcommand{\reducecaptionspace}{\vspace{-0.3cm}\xspace}
\newcommand{\reducepostfigspace}{\vspace{-0.3cm}\xspace}
\newcommand{\reducespace}{\vspace{-0.3cm}\xspace}

\newcommand{\atl}{\textsc{ATL-SN4}\xspace}
\newcommand{\dfc}{\textsc{DFC19}\xspace}

\newcommand{\flow}{\textsc{flow}\xspace}
\newcommand{\flowh}{\textsc{flow-h}\xspace}
\newcommand{\flowa}{\textsc{flow-a}\xspace}
\newcommand{\flowha}{\textsc{flow-ha}\xspace}

\newcommand{\specialcell}[2][c]{%
\begin{tabular}[#1]{@{}c@{}}#2\end{tabular}}

\newcommand{\subsubheaderbf}[1]{\mbox{\textbf{#1}\hspace*{2.5mm}}}

\title{Learning Geocentric Object Pose in Oblique Monocular Images}

\author{Gordon Christie$^{1,*}$ \hspace{0.5cm} Rodrigo Rene Rai Munoz Abujder$^{1,*}$ \hspace{0.5cm} Kevin Foster$^1$ \hspace{0.5cm} Shea Hagstrom$^1$ \hspace{0.5cm}  \\ Gregory D. Hager$^2$ \hspace{0.5cm} Myron Z. Brown$^1$ \\ 
$^1$The Johns Hopkins University Applied Physics Laboratory \quad $^2$The Johns Hopkins University \\  
{\small \tt \{\href{mailto:gordon.christie@jhuapl.edu}{gordon.christie},\href{mailto:rai.munoz@jhuapl.edu}{rai.munoz},\href{mailto:kevin.foster@jhuapl.edu}{kevin.foster}, \href{mailto:shea.hagstrom@jhuapl.edu}{shea.hagstrom}, \href{mailto:myron.brown@jhuapl.edu}{myron.brown}\}@jhuapl.edu} \\ \small \tt \href{mailto:hager@cs.jhu.edu}{hager@cs.jhu.edu} }
\maketitle

\begin{abstract}
   An object’s geocentric pose, defined as the height above ground and orientation with respect to gravity, is a powerful representation of real-world structure for object detection, segmentation, and localization tasks using RGBD images. For close-range vision tasks, height and orientation have been derived directly from stereo-computed depth and more recently from monocular depth predicted by deep networks. For long-range vision tasks such as Earth observation, depth cannot be reliably estimated with monocular images. Inspired by recent work in monocular height above ground prediction and optical flow prediction from static images, we develop an encoding of geocentric pose to address this challenge and train a deep network to compute the representation densely, supervised by publicly available airborne lidar. We exploit these attributes to rectify oblique images and remove observed object parallax to dramatically improve the accuracy of localization and to enable accurate alignment of multiple images taken from very different oblique viewpoints. We demonstrate the value of our approach by extending two large-scale public datasets for semantic segmentation in oblique satellite images. All of our data and code are publicly available\footnote{\scriptsize\url{https://github.com/pubgeo/monocular-geocentric-pose}}.
\end{abstract}

\reducespace

{\let\thefootnote\relax\footnote{{* denotes equal contribution}}}

\section{Introduction}
\label{sec:intro}

In this paper, we study the problem of rectifying oblique monocular images from overhead cameras to remove observed object parallax with respect to ground, enabling accurate object localization for Earth observation tasks including semantic mapping \cite{demir2018deepglobe}, map alignment \cite{zampieri2018multimodal,chen2019autocorrect}, change detection \cite{doshi2018satellite}, and vision-aided navigation \cite{goforth2019gps}. Current state-of-the-art methods for these tasks focus on near-nadir images without the confounding effect of parallax; however, the vast majority of overhead imagery is oblique. For response to natural disasters and other dynamic world events, often only oblique images can be made available in a timely manner. The ability to rectify oblique monocular images to remove parallax will enable a dramatic increase in utility of these methods to address real-world problems. 

\begin{figure}[t!]
    \centering
    \includegraphics[width=\columnwidth]{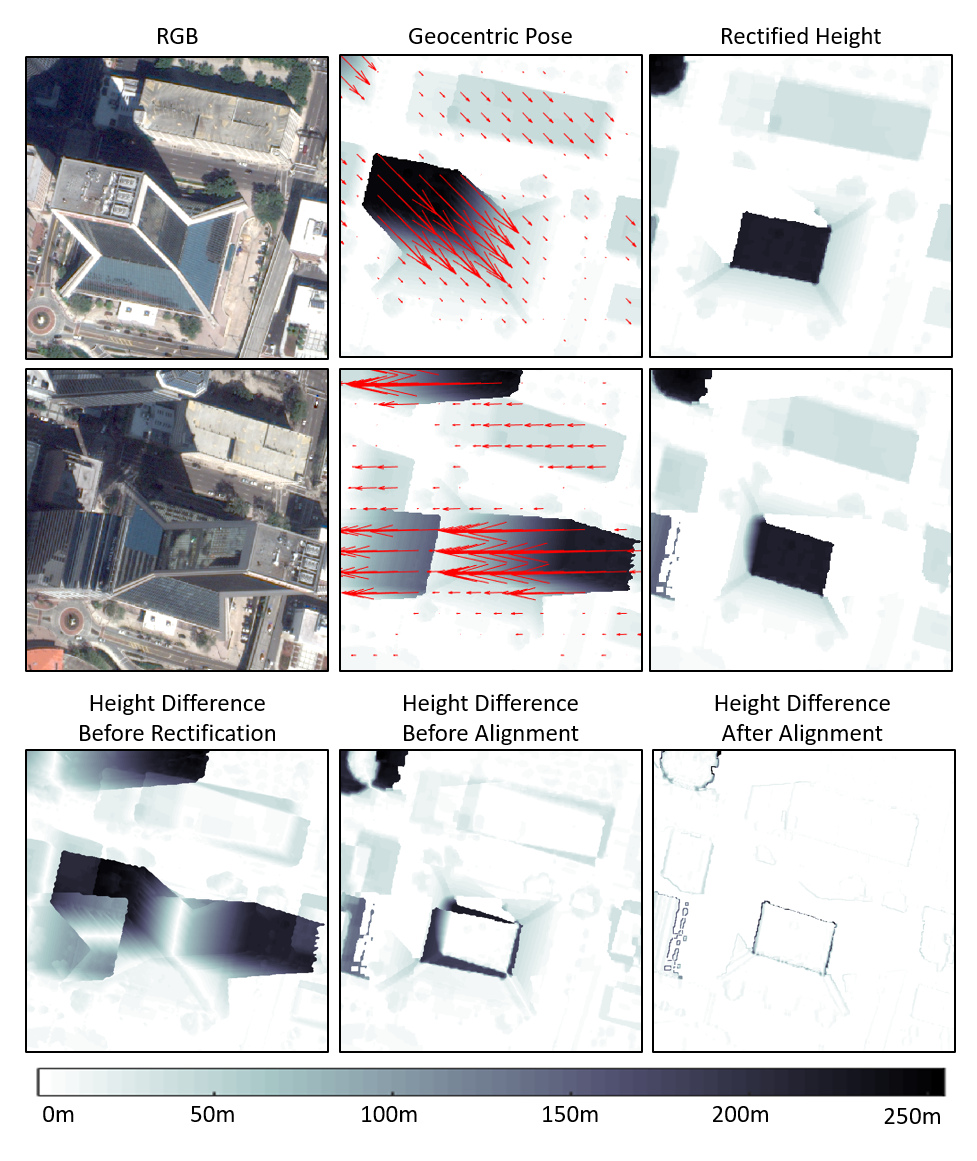}
    \reducecaptionspace
    \reducecaptionspace
    \caption{Our method takes monocular RGB images, predicts object height (meters) and geocentric pose, and rectifies height to geospatially accurate 3D models suitable for reliable alignment by a conventional method.}
    \reducepostfigspace
    \reducepostfigspace
    \label{fig:teaser}
\end{figure}

To address this very challenging problem, we first draw inspiration from Gupta et al. \cite{gupta2013perceptual} who proposed geocentric pose, or height above ground and orientation with respect to gravity, as a powerful representation to impose real-world structure on object detection, segmentation, and localization tasks using RGBD images. Hand-crafted features based on this idea have even featured prominently in state-of-the-art deep learning methods for indoor semantic segmentation \cite{gupta2014learning,long2015fully,gupta2016cross,cheng2017locality,qi20173d,lin2017cascaded,park2017rdfnet,wang2018depth}. For close-range vision tasks, height and orientation have been derived directly from stereo-computed depth and more recently from monocular depth predicted by deep networks \cite{li2018megadepth}. For long-range vision tasks such as Earth observation, depth cannot be reliably estimated with monocular images, so we further draw inspiration from recent work in monocular height above ground prediction \cite{srivastava2017joint,mou2018im2height,ghamisi2018img2dsm,amini2019cnn,bosch2019semantic,le20192019,kunwar2019u,zheng2019pop} and optical flow prediction from static images \cite{pintea2014deja,walker2015dense,walker2016uncertain,gao2018im2flow}. We develop an encoding of geocentric pose and train a deep network to compute the representation densely. Our model jointly learns to predict height above ground and dense flow vectors mapping surface features to ground level. \figref{fig:teaser} illustrates the use of our method to rectify overhead images taken from very different oblique viewpoints and then align the rectified height images – for this example, by affine homography. Height and flow for this example were derived from lidar, but example predictions from our model are shown in \secref{subsec:results}. While our experimental results are demonstrated for satellite images, we believe our method can also be successfully applied to airborne cameras and even ground-based cameras. 

Our contributions are summarized as follows:
\begin{compactitem}
\item We adopt geocentric pose as a general representation for geometry in oblique monocular images and are the first to report the following: 1) a method to supervise its learning, and 2) a method for prediction without reliance on depth estimates which cannot be reliably determined from monocular images at longer ranges.
\item We extend the Urban Semantic 3D (US3D) dataset \cite{bosch2019semantic} to include labels for the geocentric pose task, enabling public research and comparative analysis of methods. We further extend US3D to include additional images with a wide range of oblique viewing angles from the SpaceNet 4 (SN4) contest \cite{weir2019spacenet} to enable more comprehensive parametric evaluation of this task.
\item We demonstrate that our model designed to jointly learn height and orientation performs better than a model trained for each task independently, and increases efficiency through shared weights. We further demonstrate the need for rotation augmentations to overcome bias from severely limited viewpoint diversity due to sun-synchronous satellite orbits.
\item We demonstrate the efficacy of our method for image rectification to improve intersection over union (IoU) scores for semantic segmentation with oblique images.
\item All of our data and code are publicly available.
\end{compactitem}


\section{Related Work}
\label{sec:related_work}

Our approach draws inspiration from a large body of work exploiting object height and orientation with respect to ground to improve semantic segmentation and related tasks for RGBD images. Our encoding of this representation in a deep network is inspired by recent progress in predicting height above ground from single images and predicting optical flow from static images. Before introducing the details of our method, we review these motivating works.

\subsection{Geocentric Pose}
\label{subsec:geocentric_pose}
Gupta et al. \cite{gupta2013perceptual} proposed geocentric pose – height and orientation with respect to ground – as a general feature for object recognition and scene classification. Gupta et al. \cite{gupta2014learning} further proposed to encode horizontal disparity (or depth), height above ground, and orientation with respect to gravity as the popular three-channel HHA representation and demonstrated significant performance improvements for object detection, instance segmentation, and semantic segmentation tasks. Hand-crafted HHA features have since featured prominently even in deep learning state-of-the-art methods for indoor semantic segmentation \cite{long2015fully,gupta2016cross,cheng2017locality,qi20173d,lin2017cascaded,park2017rdfnet,wang2018depth} as well as object detection \cite{long2015fully, schwarz2018rgb} and semantic scene completion \cite{liu2018see}. All of these works involve close-range indoor vision tasks and derive geocentric pose from depth, with height above ground approximated relative to the lowest point in an image \cite{gupta2013perceptual}. In our work, we learn to predict these attributes directly in complex outdoor environments based on appearance without depth which is difficult to estimate reliably from images captured at long range. We also accurately predict absolute height above ground from monocular images. This is necessary for accurately rectifying the images, removing observed object parallax to improve accuracy of localization and enable accurate alignment of multiple images taken from very different oblique viewpoints.

\subsection{Monocular Height Prediction}
\label{subsec:monocular_height_prediction}
The successes of deep learning methods for monocular depth prediction \cite{li2018megadepth} have motivated recent work to directly learn to predict height from appearance in a single image. The earliest work to our knowledge was conducted by Srivastava et al. (2017) who proposed a multi-task convolutional neural network (CNN) for joint height estimation and semantic segmentation of monocular aerial images \cite{srivastava2017joint}. Mou and Zhu (2018) also proposed a CNN for height estimation and demonstrated its use for instance segmentation of buildings \cite{mou2018im2height}. Each of these early works was evaluated using a single overhead image mosaic from a single city. Ghamisi and Yokoya (2018) proposed a conditional generative adversarial network (cGAN) for image to height translation and reported results with a single image from each of three cities \cite{ghamisi2018img2dsm}. Amirkolaee and Arefi (2019) proposed a CNN trained with post-earthquake lidar and demonstrated its use to detect collapsed buildings by comparing model predictions for pre- and post-event images \cite{amini2019cnn}.
To promote research with larger-scale supervision, Bosch et al. (2019) produced the Urban Semantic 3D (US3D) dataset which includes sixty-nine satellite images over Jacksonville, FL and Omaha, NE, each covering approximately one hundred square kilometers \cite{bosch2019semantic}. Le Saux et al. (2019) leveraged this dataset to conduct the 2019 Data Fusion Contest focused on semantic 3D reconstruction, including a novel challenge track for single-view semantic 3D \cite{le20192019}. The winning solutions by Kunwar \cite{kunwar2019u} and Zheng et al. \cite{zheng2019pop} both exploited semantic labels as priors for height prediction. In this work, we demonstrate comparable accuracy without semantic priors. We also show improved height prediction accuracy by jointly learning to predict orientation flow vectors. In addition to our experiments, we leverage and extend the US3D dataset using public satellite images from the 2018 SpaceNet 4 (SN4) contest that span a wide range of viewing angles over Atlanta, GA \cite{weir2019spacenet}, and we demonstrate that our method to predict geocentric pose significantly improves building segmentation accuracy for oblique images.

\subsection{Optical Flow Prediction from a Static Image}
\label{subsec:optical_flow_prediction_static_image}
Our approach to learning geocentric pose is inspired by recently demonstrated methods to predict dense optical flow fields from static images with self-supervision from optical flow methods applied to videos. Pintea et al. (2014) proposed regression of dense optical flow fields from static images using structured random forests \cite{pintea2014deja}. Walker (2015) proposed a CNN for ordinal regression to better generalize over diverse domains \cite{walker2015dense}. Walker et al. (2016) proposed a generative model using a variational auto-encoder (VAE) for learning motion trajectories from static images \cite{walker2016uncertain}. Gao et al. (2018) also explored a generative model using a cGAN but reported state-of-the-art results for optical flow prediction and action recognition with their Im2Flow regression model, a modified U-Net CNN encoder/decoder trained by minimizing both a pixel L2 loss and a motion content loss derived from a separate action recognition network that regularizes the regression network to produce realistic motion patterns \cite{gao2018im2flow}. To learn geocentric pose, we employ a similar U-Net architecture and demonstrate improved performance by jointly learning to predict height. We also highlight orientation bias for our task by performing rotation augmentations during training. We produce reference flow fields for supervision automatically using lidar as discussed in \secref{subsec:supervision}.


\section{Learning Geocentric Pose}
\label{sec:learning_geocentric_pose}

\begin{figure*}[t!]
    \centering
    \includegraphics[width=\textwidth]{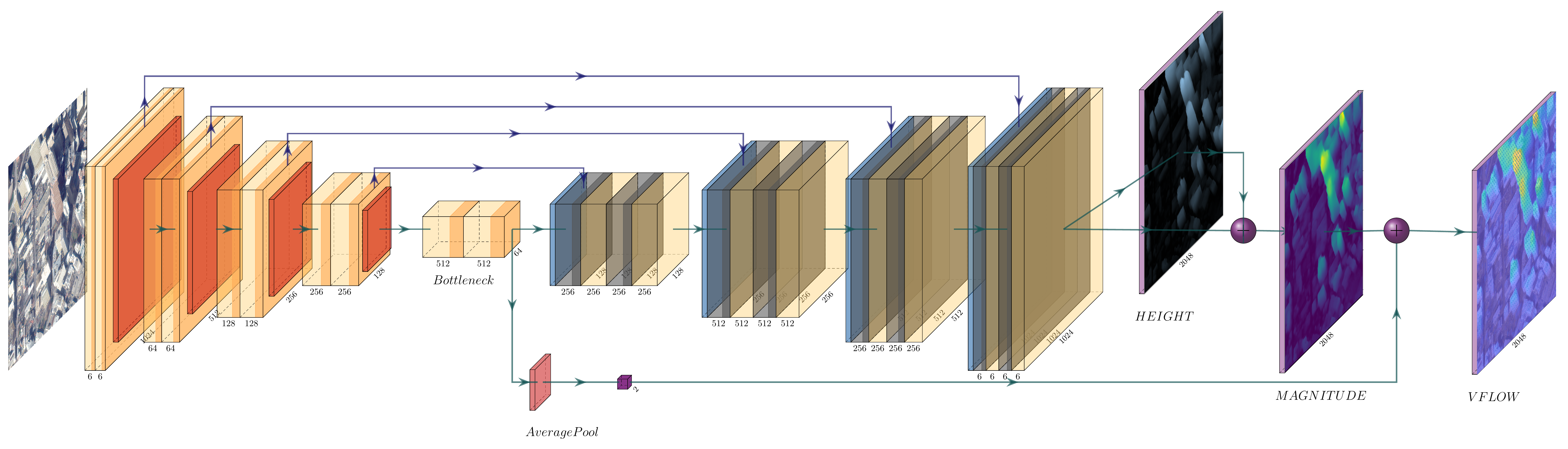}
    \reducecaptionspace
    \caption{This shows the architecture of our full approach, which uses a U-Net decoder with a ResNet34 encoder. At the last layer of the encoder, we predict the image-level orientation as $\protect\sin(\protect\theta)$ and $\protect\cos(\protect\theta)$. At the output of the decoder, we predict per-pixel above-ground-level height values, which are concatenated with the decoder's output and used to predict per-pixel magnitudes. An MSE loss is used for all output heads. At test time, flow vectors can be calculated by multiplying image-level orientation predictions with the per-pixel magnitudes.}
    \reducepostfigspace
    \label{fig:architecture}
\end{figure*}

\subsection{Representation}
\label{subsec:representation}
Our representation of geocentric pose encodes height above ground and flow vectors that map surface features to ground level. A satellite pushbroom sensor model is well-approximated locally by affine projection which preserves the invariant property of parallelism \cite{de2014stereo}. We exploit this property in representing flow fields with pixel-level magnitudes and image-level orientation. Similar to \cite{gao2018im2flow}, we represent orientation ($\theta$) as a two-element vector, [$\sin(\theta)$, $\cos(\theta)$], representing the horizontal and vertical components of the flow vectors. We observe that each feature's height above ground is intrinsic and the magnitude of its flow vector is related to that height by each image's projection. We thus employ height as a prior in our model for learning magnitude.

\subsection{Model}
\label{subsec:model}
Our model, illustrated in \figref{fig:architecture}, jointly predicts image-level orientation, as well as dense above-ground-level heights and flow vector magnitudes. The base architecture utilizes a U-Net decoder with a ResNet34 encoder. At the last layer of the encoder, the image-level orientation is predicted as $\sin(\theta)$ and $\cos(\theta)$. The output of the decoder is used to predict heights, which are concatenated with the decoder output for predicting magnitudes. MSE is used for all output heads (image-level orientation, magnitude, and height), where each loss is weighted equally during training. At test time, flow vectors can be calculated by multiplying the predictions of image-level orientation and per-pixel magnitudes.
We present an ablation study where height prediction is removed from the model to show its importance for learning to predict orientation and magnitude. Height is intrinsic to objects in the image, where pixels representing the same physical locations on a building in different images should have the same heights. However, magnitudes for these pixels will vary with changes to viewing geometry. We believe the intrinsic properties of height provide valuable context for predicting magnitude. We also show that the accuracy of our height predictions is comparable to state-of-the-art solutions for a public challenge dataset, and note that our network shares weights for multiple tasks, making it more efficient than having separate networks for each task.  	
	
\subsection{Supervision}
\label{subsec:supervision}
\begin{figure}[t!]
    \centering
    \includegraphics[width=\columnwidth]{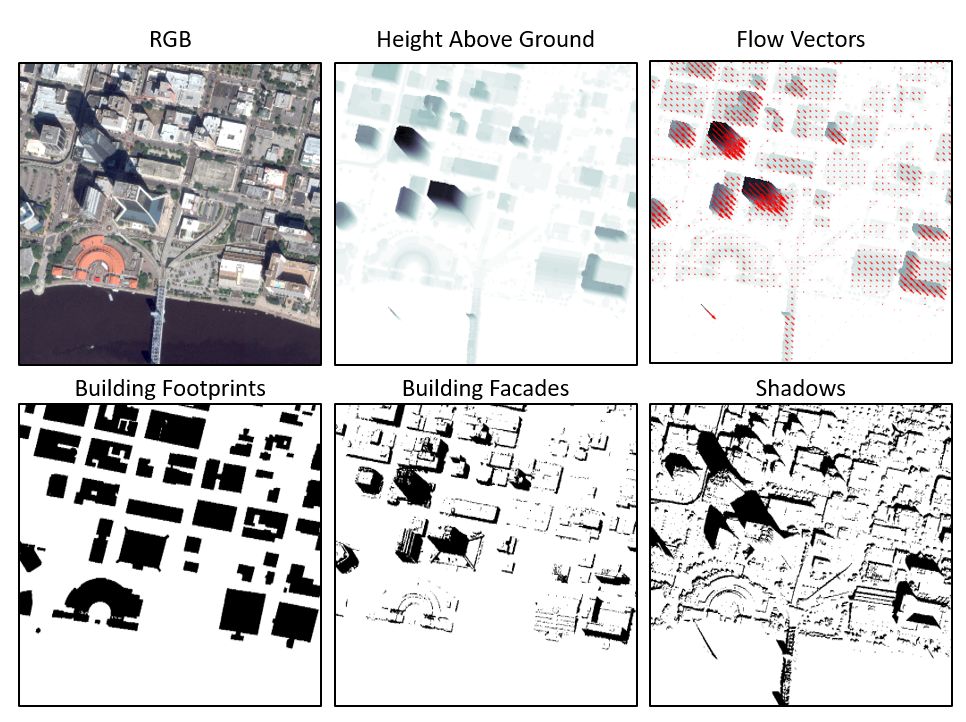}
    \reducecaptionspace
    \caption{Lidar-derived attributes for each RGB image include height above ground, geocentric pose flow vectors, and shadow masks. Map attributes include semantic labels, building facades, and ground-level building footprints.}
    \reducepostfigspace
    \label{fig:lidar_attributes}
\end{figure}

To enable supervised learning of our model, we have developed a pipeline for producing non-overlapping overhead RGB image tiles with lidar-derived attributes projected into each oblique image pixel, as illustrated in \figref{fig:lidar_attributes}. We utilized this pipeline to produce training and test datasets for our task, augmenting public data from US3D \cite{bosch2019semantic} and SN4 \cite{weir2019spacenet}. For each geographic tile, we first align each overhead image with lidar intensity using the mutual information metric and update the image translation terms in the RPC camera metadata \cite{de2014stereo}. To improve reliability of image matching, we cast shadows in each lidar intensity image using solar angle image metadata to match the shadows observed in the RGB image. Layers produced include UTM geographic coordinates, ground-level height from the Digital Terrain Model (DTM), surface-level height from the Digital Surface Model (DSM), height above ground computed from the difference of the DSM and DTM, the shadow mask produced for image matching, and image flow vectors mapping surface-level feature pixels to their ground-level pixel coordinates. Our representation of geocentric pose is composed of height above ground and orientation with respect to ground as defined by the dense flow vectors. Both rely on knowledge of ground level in the DTM. For the lidar data used in our experiments, DTM layers were produced by professional surveyors with manual edits, but automated methods for ground classification in lidar and even in DSMs produced using satellite images also work well \cite{duan2019large}.

For our experiments, we also employ semantic labels derived from public map data to demonstrate the value of our model for rectifying map features in oblique images. We project this map data into each image with the same procedure used for lidar attributes. Layers include semantic label for each pixel and ground-level footprints for buildings. Building facades are labeled separately from roofs.

\section{Experiments}
\label{sec:experiments}

\subsection{Datasets}
\label{subsec:datasets}
For our experiments, we extended two publicly-available datasets – US3D \cite{bosch2019semantic} and SN4 \cite{weir2019spacenet} – using the method described in \secref{subsec:supervision} and illustrated in \figref{fig:lidar_attributes}. We train with the full resolution for each dataset.

\begin{figure}[t!]
    \centering
    \includegraphics[width=\columnwidth]{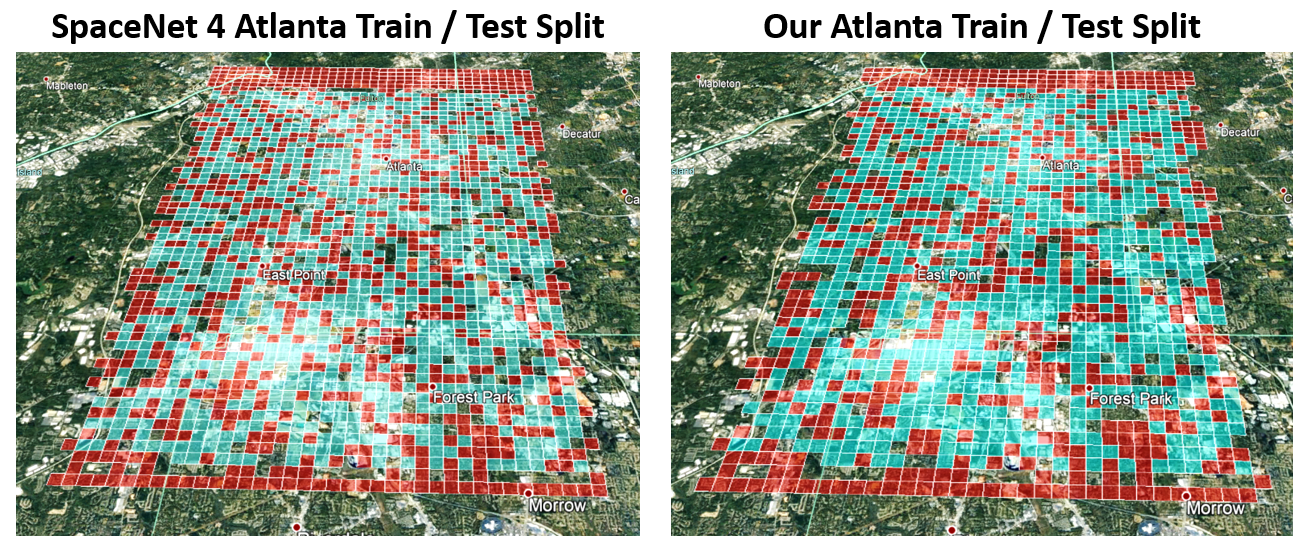}
    \reducecaptionspace
    \caption{Train (blue) and test (red) tiles for the \atl unrectified images (right) were selected to closely match the split for SpaceNet 4 orthorectified image tiles (left). Images shown are from Google Earth.}
    \reducepostfigspace
    \label{fig:atl_sn4_tiles}
\end{figure}

\begin{compactitem}
\item \dfc. We use the same 2,783 training images and 50 testing images of Jacksonville, FL and Omaha, NE from US3D used for the 2019 Data Fusion Contest \cite{le20192019}. We also use an extended test set with 300 images including more view diversity for the same geographic tiles. Images are each 2048x2048 pixels.
\item \atl. We produced 25,500 training images and 17,554 testing images of Atlanta, GA using public unrectified source images to closely match the rectified image tiles used for SN4, as shown in \figref{fig:atl_sn4_tiles}. We used 7,702 training images and 310 testing images, cropped to 1024x1024 pixels, for our experiments. 
\end{compactitem}

\begin{figure}[t!]
    \centering
    \includegraphics[width=\columnwidth]{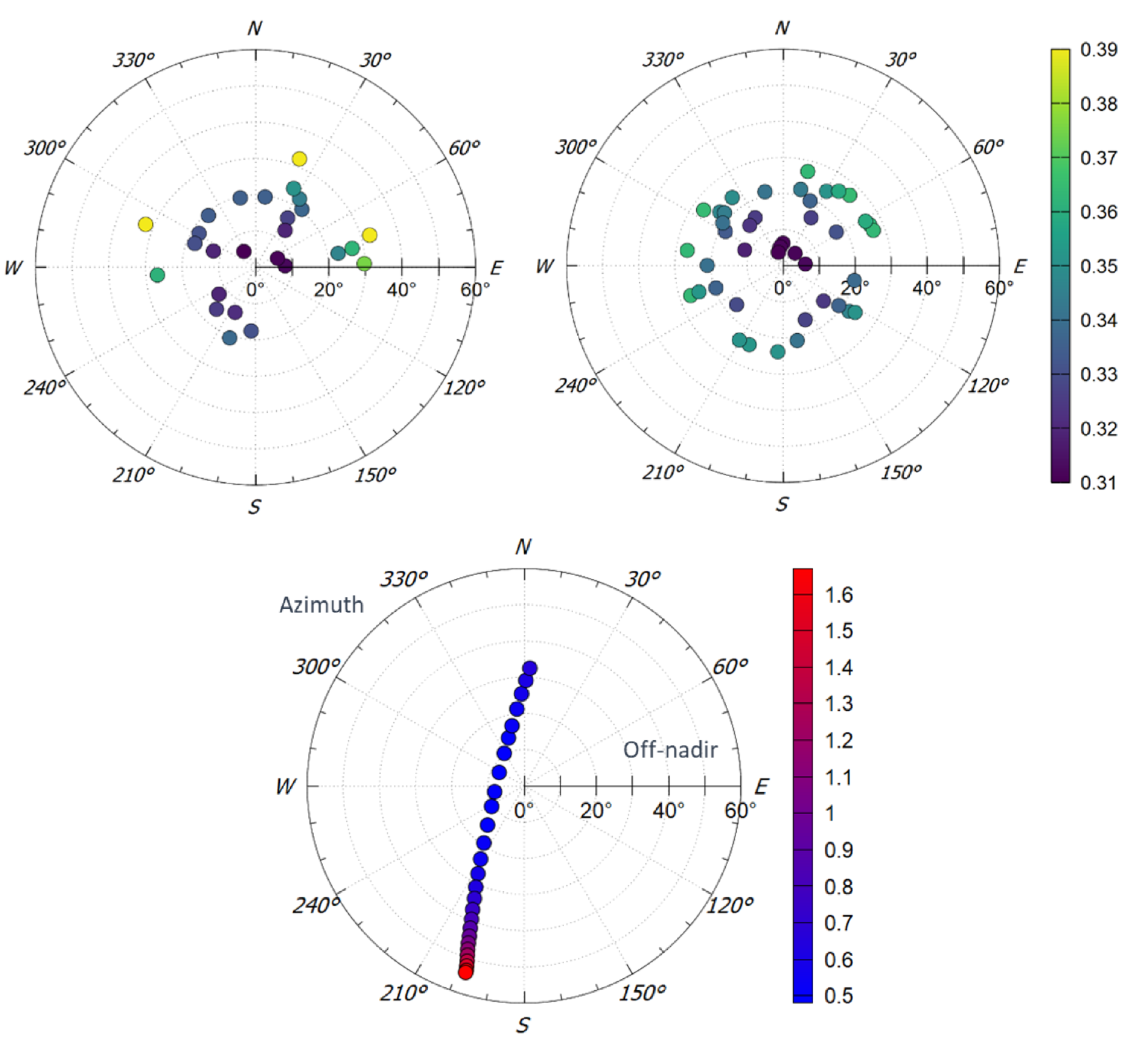}
    \reducecaptionspace
    \caption{Azimuth angle, off-nadir angle, and resolution (meters) is shown for Jacksonville and Omaha images from \dfc (top) and \atl images (bottom).}
    \reducepostfigspace
    \label{fig:angles_distribution}
\end{figure}

Viewpoint diversity and pixel resolution for images in the \dfc and \atl datasets are shown in \figref{fig:angles_distribution}. Jacksonville and Omaha images were collected by MAXAR’s WorldView-3 satellite on multiple dates with a variety of azimuth angles and limited off-nadir angles. \atl images were collected by MAXAR’s WorldView-2 satellite during a single orbit with very limited azimuth diversity and a wide range of off-nadir angles. Together, these datasets enable thorough evaluation.

\subsection{Methods}
\label{subsec:methods}
\paragraph{Flow Vector Regression}
For each test set, we present four sets of results. These include combinations of models trained with and without height supervision, and with and without train-time rotation augmentations. As discussed earlier, our datasets consist of orientation bias because of the sun-synchronous satellite orbits. To make our model generalizable to unseen orientations, we perform train-time flips and rotations randomly, which can introduce new orientation ground truth for each image at different epochs during training. Our approaches are described as follows:

\begin{compactitem}
\item \subsubheaderbf{\flow} Model with height prediction head removed and trained without augmentations. 
\item \subsubheaderbf{\flowh} Full model trained without augmentations.
\item \subsubheaderbf{\flowa} \flow trained with augmentations.
\item \subsubheaderbf{\flowha} \flowh trained with augmentations.
\end{compactitem}

For completeness, we present image-level orientation (angle) and pixel-level magnitude (mag) errors for our predictions, as they are learned separately during training. Orientation errors are measured in degrees, while magnitude errors are measured in pixels. However, we note that orientation and magnitude are typically not appropriate metrics for this task. As an example, in a nadir image where all pixel magnitudes are zero, predicting the orientation is meaningless. Similarly, in a highly-oblique image where the magnitudes are high, it is extremely important to predict the orientation accurately. We therefore measure per-pixel endpoint errors (EPE), which measure the Euclidean distance between the endpoints of the predicted and ground truth flow vectors. However, note that mag errors are equal to EPE when orientation is known from the sensor metadata, which is sometimes the case with satellite imagery. Therefore, mag errors can be an appropriate metric when orientation is known. 

These metrics are calculated with and without test-time rotations to show how models that do not include train-time rotations over-fit to the limited set of orientations in the train set. We also calculate per-category EPE to show how semantics affect performance. Categories from \dfc are used, as well as a separate layer with shadow masks.

\paragraph{Building Footprint Extraction}
One of the goals of this work is to enable more accurate automated-mapping from overhead imagery. With our flow vector predictions, outputs from any segmenter or detector can be input into our model and transformed to ground level. To demonstrate the accuracy of our model, we use building annotations and footprints from the \dfc and \atl test sets. Building annotations consist of the roof and facade labels in the image, while the footprints represent the base of the building identified from top-down lidar. Using our predicted flow vectors, we warp the building annotations to ground level and compare to the ground truth footprints. 

We also demonstrate the reverse capability, where we start with footprints and warp them into building annotations using our predicted flow vectors. This is useful in situations where there is a desire to overlay map data (e.g., OpenStreetMap) on imagery as an initial set of annotations. For example, when a new image is captured of an area actively being developed, we may want to pull in existing annotations so annotators do not start from scratch.

We compare three results for each of the two tasks: 1) transform building annotations to footprints, and 2) transform footprints to building annotations. First, we measure IoU between the building annotations and the footprints to understand what the accuracy is when we do nothing. Second, we warp the source mask (building annotations or footprints) to the target mask using the ground truth flow vectors to get an upper bound for the IoU on what can be achieved if we perfectly predict the flow vectors. Note that we do not get perfect overlap in this case because of occluded ground pixels. Finally, we measure IoU for the warped versions of the source masks using our predicted flow vectors. 

\subsection{Results}
\label{subsec:results}
\paragraph{Height Prediction}
We assess our current method, which takes the height outputs of \flowh, compared to two recent strong baselines \cite{kunwar2019u,zheng2019pop} for the very challenging \dfc test set \cite{le20192019}, measuring mean and root mean square (RMS) error (meters) for height predictions compared to above ground height measured from lidar. Results are shown in \tabref{tab:height_pred_acc}. Both baseline methods anchor height predictions using semantic category, and both exploit test-time ensembles to improve performance. While semantic anchors appear to improve accuracy for categories with low height variance, they do not account for the variance observed in urban scenes. Our model performs better overall without semantic priors or test-time ensembles. 

\begin{figure}[t!]
    \centering
    \includegraphics[width=\columnwidth]{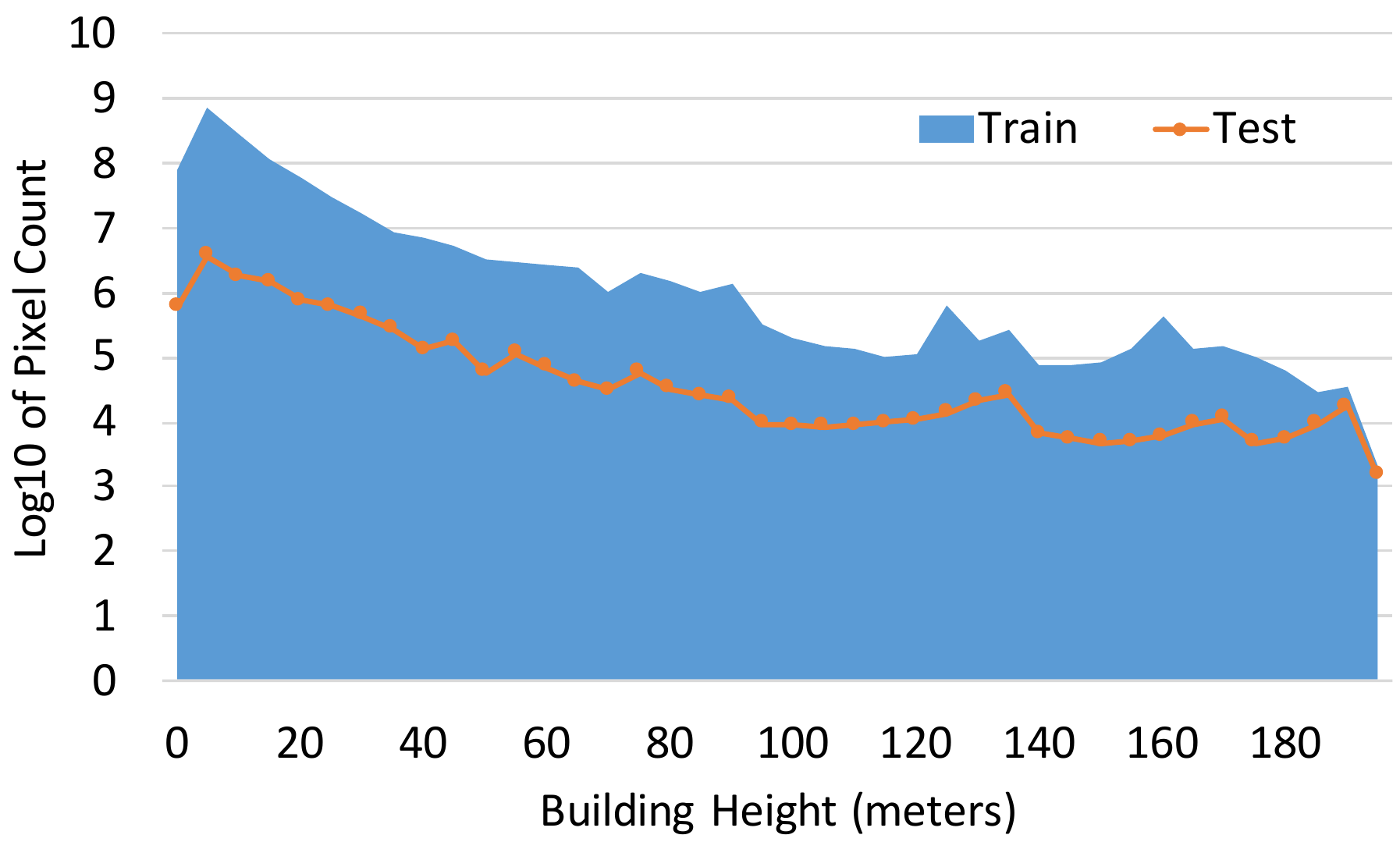}
    \reducecaptionspace
    \caption{Height distributions in train and test sets are comparable, with some buildings approaching 200 meters.}
    \reducepostfigspace
    \label{fig:height_distributions}
\end{figure}

\begin{figure}[t!]
    \centering
    \includegraphics[width=\columnwidth]{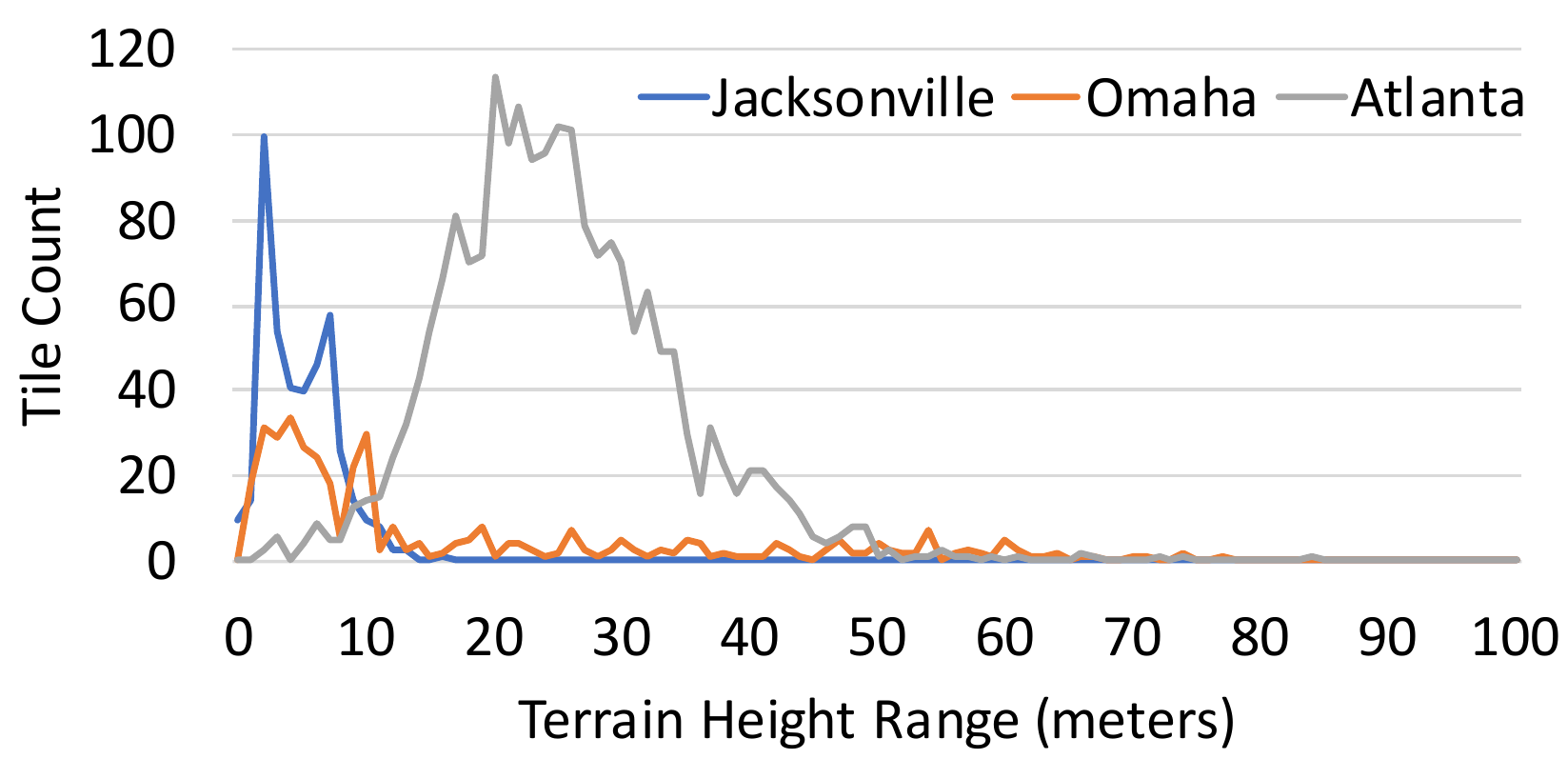}
    \reducecaptionspace
    \caption{Ground-level terrain height variation statistics.}
    \reducepostfigspace
    \label{fig:ground_level_terrain_height_var}
\end{figure}

\figref{fig:height_distributions} depicts building height statistics for the train and test sets, with some building heights approaching 200 meters. Achieving more reliable predictions for those rare tall objects is a topic for ongoing research. Height prediction performance in the presence of significant terrain relief has also yet to be characterized. Statistics for ground-level terrain height variation in the \dfc and \atl data sets are shown in \figref{fig:ground_level_terrain_height_var}.

\begin{table}[h!]
	\centering
	\setlength{\tabcolsep}{5pt}
	{\small
		\scalebox{0.9}{
			\begin{tabular}{ccccc}
				\toprule  
				 & \textbf{mean} & \textbf{mean bldgs} & \textbf{RMS} & \textbf{RMS bldgs} \\
				\midrule
				Kunwar \cite{kunwar2019u} & \textbf{2.69} & 8.33 & 9.26 & 19.65 \\
				Zheng et al. \cite{zheng2019pop} & 2.94 & 8.72& 9.24 & 19.32 \\
				Ours & 2.98 & \textbf{7.73} & \textbf{8.23} & \textbf{16.87} \\
				\bottomrule
			\end{tabular}
		}
	}
    \reducecaptionspace
	\caption{Our regression model produces height predictions with lower RMS error (meters) than baseline models that anchor height predictions with semantic category.}
    \reducepostfigspace
	\label{tab:height_pred_acc}
\end{table}

\paragraph{Flow Vector Regression}
Our results for each of the approaches on the \dfc test set without test-time augmentations can be seen in \tabref{tab:dfc19_no_test_rot}. The results from the same approaches applied to the test set containing rotation augmentations are shown in \tabref{tab:dfc19_with_test_rot}. The per-category results are EPE. Results in shadows, which are a separate layer (i.e., not included as part of the DFC category layer) are also included. \tabref{tab:atl_without_rot} and \tabref{tab:atl_with_rot} show similar results for \atl, but exclude a semantic breakdown, as the same human-validated semantic labels are not available for this dataset. The test sets consist of the original \dfc and \atl test sets along with 9 additional rotation augmentations per image at intervals of 36 degrees. 

Two key observations can be made about these results. 1) It is clear from \tabref{tab:dfc19_with_test_rot} and \tabref{tab:atl_with_rot} that models trained without rotation augmentations over-fit to the orientation bias of the train set, and that train-time rotation augmentations are currently needed to create generalizable models for this task. 2) Jointly learning to predict above-ground-height improves metrics across most categories when test-time rotations are applied. Unsurprisingly, we observe the lowest EPE values for ground pixels, and some of the highest EPE errors on facades, roofs, and elevated roads, where ground truth magnitudes are highest.

\begin{figure}[t!]
    \centering
    \includegraphics[width=\columnwidth]{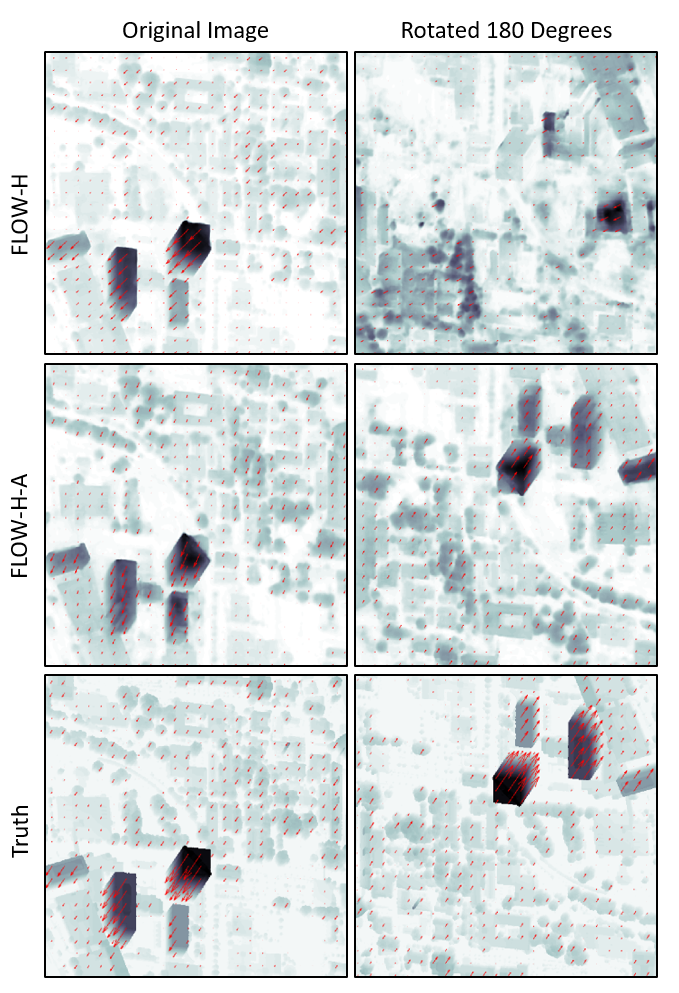}
    \reducecaptionspace
    \caption{Orientation augmentations in training our model help to reduce bias in the satellite viewing angles. Height and flow vector ground truth and predictions from models trained with and without augmentations are shown for an example from \atl.}
    \reducepostfigspace
    \label{fig:qualitative_aug_result}
\end{figure}

\begin{table*}[ht!]
	\centering
	\setlength{\tabcolsep}{8pt}
    \begin{tabular}{c|ccc|cccccc|c}
        \toprule  
        \textbf{Method} & \textbf{mag} & \textbf{angle} & \textbf{EPE} & \textbf{ground} & \textbf{veg} & \textbf{roof} & \textbf{water} & \textbf{elevated roads} & \textbf{facade} & \textbf{shadow} \\
        \midrule
        \flow & 2.71 & 16.11 & 3.08 & 1.39 & \textbf{3.68} & \textbf{5.44} & 1.78 & 6.86 & \textbf{7.11} & \textbf{4.03} \\
        \flowh & \textbf{2.40} & 16.14 & \textbf{2.92} & \textbf{0.92} & 3.86 & 5.70 & 1.54 & \textbf{6.42} & 7.37 & 3.98 \\
        \flowa & 2.91 & 17.52 & 3.24 & 1.15 & 4.04 & 6.17 & 1.57 & 7.66 & 8.32 & 4.42 \\
        \flowha & 2.69 & \textbf{15.09} & 3.04 & 1.06 & 4.06 & 5.89 & \textbf{1.41} & 6.89 & 7.83 & 4.25 \\
        \bottomrule
    \end{tabular}
	\caption{Results \textbf{without} test-time rotations for \dfc. Lower is better for all numbers. Per-category values are all end point errors (EPE). This table highlights that models trained to generalize perform worse than models that learn the orientation bias of the train set. However, we note that the model trained without rotation augmentations and with height supervision has the best overall EPE.}
	\label{tab:dfc19_no_test_rot}
\end{table*}

\begin{table*}[ht!]
	\centering
	\setlength{\tabcolsep}{8pt}
    \begin{tabular}{c|ccc|cccccc|c}
        \toprule  
        \textbf{Method} & \textbf{mag} & \textbf{angle} & \textbf{EPE} & \textbf{ground} & \textbf{veg} & \textbf{roof} & \textbf{water} & \textbf{elevated roads} & \textbf{facade} & \textbf{shadow} \\
        \midrule
        \flow & 4.15 & 79.52 & 6.11 & 2.39 & 7.34 & 11.99 & 3.01 & 12.67 & 13.80 & 7.50 \\
        \flowh & 4.07 & 78.15 & 5.95 & 2.06 & 7.29 & 12.18 & 2.94 & 12.82 & 13.86 & 7.35 \\
        \flowa & 3.02 & 17.48 & 3.35 & 1.18 & \textbf{4.12} & 6.22 & 1.56 & 8.06 & 8.35 & 4.51 \\
        \flowha & \textbf{2.83} & \textbf{16.79} & \textbf{3.21} & \textbf{1.10} & 4.17 & \textbf{6.10} & \textbf{1.44} & \textbf{7.55} & \textbf{8.08} & \textbf{4.42} \\
        \bottomrule
    \end{tabular}
	\caption{Results \textbf{with} test-time rotations for \dfc. Lower is better for all numbers. Per-category values are all end point errors (EPE). This table highlights that train-time rotation augmentations are currently needed to overcome the orientation bias caused by sun-synchronous satellite orbits and perform well in the presence of test-time rotations. These results also highlight that training with height supervision improves overall EPE performance over most categories.  These improvements are most notable for roof, elevated roads, and facades, where accurate flow vector prediction is more important. }
	\label{tab:dfc19_with_test_rot}
\end{table*}

We show the importance of train-time rotations qualitatively in \figref{fig:qualitative_aug_result}. In the first column, where no test-time rotation was performed, we can qualitatively observe similar performance between \flowh and \flowha. However, in the second column, when we rotate the image to an orientation not originally represented in the train set, we see \flowh qualitatively performing worse than \flowha.

\begin{table}[h!]
	\centering
	\setlength{\tabcolsep}{16pt}
    \begin{tabular}{cccc}
        \toprule  
        \textbf{Method} & \textbf{mag} & \textbf{angle} & \textbf{EPE} \\
        \midrule
        \flow & 3.88 & 9.64 & 4.17 \\
        \flowh & \textbf{3.78} & \textbf{7.38} & \textbf{3.99} \\
        \flowa & 5.37 & 15.76 & 6.03 \\
        \flowha & 4.79 & 16.57 & 5.38 \\
        \bottomrule
    \end{tabular}
	\caption{Results \textbf{without} test-time rotations for \atl. Similar to \tabref{tab:dfc19_no_test_rot}, we see that the model trained without rotation augmentations, but with height supervision, performs best when the test set contains orientation bias.}
    \reducepostfigspace
	\label{tab:atl_without_rot}
\end{table}

\begin{table}[h!]
	\centering
	\setlength{\tabcolsep}{16pt}
    \begin{tabular}{cccc}
        \toprule  
        \textbf{Method} & \textbf{mag} & \textbf{angle} & \textbf{EPE} \\
        \midrule
        \flow & 6.04 & 77.31 & 8.79 \\
        \flowh & 6.30 & 81.34 & 9.04 \\
        \flowa & 4.81 & \textbf{15.77} & 5.39 \\
        \flowha & \textbf{4.22} & 23.19 & \textbf{5.15} \\
        \bottomrule
    \end{tabular}
	\caption{Results \textbf{with} test-time rotations for \atl. Similar to \tabref{tab:dfc19_with_test_rot}, we see that train-time rotations and height supervision are important when test-time rotations are applied.}
    \reducepostfigspace
	\label{tab:atl_with_rot}
\end{table}

\paragraph{Building Footprint Extraction}
In this section, we demonstrate the ability to transform semantic segmentations in the image space to ground-level map data, as well as pulling map data into imagery. \tabref{tab:iou_bldg_dfc} and \tabref{tab:iou_bldg_atl} show IoU for \dfc and \atl, respectively. Unrectified is the comparison between the building annotations and the footprints without warping. Ours is the comparison between warped versions of the original mask and target mask using the predicted flow vectors. GT follows the same process as Ours, but with the ground truth flow vectors.

As seen from \tabref{tab:iou_bldg_dfc} and \tabref{tab:iou_bldg_atl}, our results better capture the footprints in these datasets than the original building annotations. Note that occluded pixels prevent GT from reaching an IoU score of 1. GT represents an upper bound on what can be achieved with perfect flow vector prediction.

\begin{table}[h!]
	\centering
	\setlength{\tabcolsep}{16pt}
    \begin{tabular}{ccc}
        \toprule  
         & \textbf{\specialcell{Building to \\ Footprint}} & \textbf{\specialcell{Footprint to \\ Building}} \\
        \midrule
        Unrectified & 0.78 (92.9\%) & 0.78 (90.7\%) \\
        Ours & 0.83 (98.8\%) & 0.82 (95.3\%) \\
        GT & 0.84 & 0.86 \\
        \bottomrule
    \end{tabular}
    \reducecaptionspace
	\caption{IoU and percentage of GT for transforming building annotations to footprints and vice versa for \dfc. }
	\reducepostfigspace
    \label{tab:iou_bldg_dfc}
\end{table}

\begin{table}[h!]
	\centering
	\setlength{\tabcolsep}{16pt}
    \begin{tabular}{ccc}
        \toprule  
         & \textbf{\specialcell{Building to \\ Footprint}} & \textbf{\specialcell{Footprint to \\ Building}} \\
        \midrule
        Unrectified & 0.74 (89.2\%)& 0.74 (86.0\%) \\
        Ours & 0.76 (91.6\%) & 0.77 (89.5\%) \\
        GT & 0.83 & 0.86 \\
        \bottomrule
    \end{tabular}
    \reducecaptionspace
	\caption{IoU and percentage of GT for transforming building annotations to footprints and vice versa for \atl. }
	\reducepostfigspace
    \label{tab:iou_bldg_atl}
\end{table}

\paragraph{Map Alignment}
Rectifying semantic labels to ground level simplifies the task of aligning maps and oblique images as shown in \figref{fig:teaser}. To demonstrate this, we apply the MATLAB imregdemons function, an efficient implementation of non-parametric image registration \cite{vercauteren2009diffeomorphic}, to estimate dense displacement fields between pairs of images in the \dfc test set. We do this for aligning RGB images as a baseline and then for rectified height images to demonstrate improved alignment. \tabref{tab:iou_overlapping_images} shows mean IoU scores for reference building segmentation labels rectified to ground level and compared with the reference footprints after alignment. Mean IoU is significantly improved, and the fraction of images with IoU greater than 0.5 is significantly improved.

\begin{table}[h!]
	\centering
	\setlength{\tabcolsep}{14pt}
    \begin{tabular}{ccc}
        \toprule  
         & \textbf{Mean} & \textbf{IoU $>$ 0.5} \\
        \midrule
        Unaligned & 0.46 & 0.40 \\
        RGB aligned & 0.66 & 0.85 \\
        \flowha & 0.69 & 0.93 \\
        \flowha fixed angle & \textbf{0.69} & \textbf{0.94} \\
        \bottomrule
    \end{tabular}
	\caption{IoU values for transforming per-pixel building annotations to footprints in other overlapping images.}
	\label{tab:iou_overlapping_images}
\end{table}


\section{Discussion}
\label{sec:discussion}

\vspace{-0.2cm}

In this paper, we have introduced the novel task of learning geocentric pose, defined as height above ground and orientation with respect to gravity, for above-ground objects in oblique monocular images. While we have shown the value of this representation for rectifying above-ground features in oblique satellite images, we believe that with minor modifications our method can also be successfully applied to airborne cameras and even ground-based cameras to address a broad range of outdoor mapping, change detection, and vision-aided navigation tasks for which a single ground plane cannot be assumed.

Much of the prior work on geocentric pose has focused on its exploitation as hand-crafted features for semantic segmentation. In this work, we have focused on its exploitation to rectify building segmentations to ground level, enabling geospatially accurate mapping with oblique images. Similar to much prior work with the HHA representation, we expect that our representation will also provide an effective prior for regularizing semantic segmentation predictions.

While our current results clearly indicate the efficacy of the proposed method, much remains unexplored. We expect that more explicitly employing intuitive cues such as shadows and building facades will help reduce prediction error for the height variation observed in urban scenes. Further, while our rotation augmentations help account for orientation bias in satellite images, we expect that more fully accounting for true geometry and appearance variation will help address current observed failure cases. We plan to explore these ideas in future work, and we will publicly release all of our code and data.

\section*{Acknowledgments}
This work was supported by the Intelligence Advanced Research Projects Activity (IARPA) contract no. 2017-17032700004. This work was further supported by the National Geospatial-Intelligence Agency (NGA) and approved for public release, 20-316, with distribution statement A – approved for public release; distribution is unlimited. The U.S. Government is authorized to reproduce and distribute reprints for Governmental purposes notwithstanding any copyright annotation thereon. Disclaimer: The views and conclusions contained herein are those of the authors and should not be interpreted as necessarily representing the official policies or endorsements, either expressed or implied, of IARPA, NGA, or the U.S. Government.

\clearpage

{\small
\bibliographystyle{ieee_fullname}
\bibliography{references}

\begin{thebibliography}{10}\itemsep=-1pt

\bibitem{amini2019cnn}
Hamed Amini~Amirkolaee and Hossein Arefi.
\newblock {CNN-based estimation of pre-and post-earthquake height models from
  single optical images for identification of collapsed buildings}.
\newblock {\em Remote Sensing Letters}, 2019.

\bibitem{bosch2019semantic}
Marc Bosch, Kevin Foster, Gordon Christie, Sean Wang, Gregory~D Hager, and
  Myron Brown.
\newblock {Semantic Stereo for Incidental Satellite Images}.
\newblock In {\em WACV}, 2019.

\bibitem{chen2019autocorrect}
Honglie Chen, Weidi Xie, Andrea Vedaldi, and Andrew Zisserman.
\newblock {AutoCorrect: Deep Inductive Alignment of Noisy Geometric
  Annotations}.
\newblock {\em BMVC}, 2019.

\bibitem{cheng2017locality}
Yanhua Cheng, Rui Cai, Zhiwei Li, Xin Zhao, and Kaiqi Huang.
\newblock {Locality-Sensitive Deconvolution Networks with Gated Fusionfor RGB-D
  Indoor Semantic Segmentation}.
\newblock In {\em CVPR}, 2017.

\bibitem{de2014stereo}
Carlo de Franchis, Enric Meinhardt-Llopis, Julien Michel, J-M Morel, and
  Gabriele Facciolo.
\newblock On stereo-rectification of pushbroom images.
\newblock In {\em ICIP}, 2014.

\bibitem{demir2018deepglobe}
Ilke Demir, Krzysztof Koperski, David Lindenbaum, Guan Pang, Jing Huang, Saikat
  Basu, Forest Hughes, Devis Tuia, and Ramesh Raska.
\newblock {DeepGlobe 2018: A Challenge to Parse the Earth through Satellite
  Images}.
\newblock In {\em CVPRW}, 2018.

\bibitem{doshi2018satellite}
Jigar Doshi, Saikat Basu, and Guan Pang.
\newblock {From Satellite Imagery to Disaster Insights}.
\newblock {\em NeurIPS Workshops}, 2018.

\bibitem{duan2019large}
Liuyun Duan, Mathieu Desbrun, Anne Giraud, Fr{\'e}d{\'e}ric Trastour, and
  Lionel Laurore.
\newblock {Large-Scale DTM Generation From Satellite Data}.
\newblock In {\em CVPRW}, 2019.

\bibitem{gao2018im2flow}
Ruohan Gao, Bo Xiong, and Kristen Grauman.
\newblock {Im2Flow: Motion Hallucination from Static Images for Action
  Recognition}.
\newblock In {\em CVPR}, 2018.

\bibitem{ghamisi2018img2dsm}
Pedram Ghamisi and Naoto Yokoya.
\newblock {IMG2DSM: Height Simulation From Single ImageryUsing Conditional
  Generative Adversarial Net}.
\newblock {\em IEEE Geoscience and Remote Sensing Letters}, 2018.

\bibitem{goforth2019gps}
Hunter Goforth and Simon Lucey.
\newblock {GPS-Denied UAV Localization using Pre-existing Satellite Imagery}.
\newblock In {\em ICRA}, 2019.

\bibitem{gupta2013perceptual}
Saurabh Gupta, Pablo Arbelaez, and Jitendra Malik.
\newblock {Perceptual organization and recognition of indoor scenes from RGB-D
  images}.
\newblock In {\em CVPR}, 2013.

\bibitem{gupta2014learning}
Saurabh Gupta, Ross Girshick, Pablo Arbelaez, and Jitendra Malik.
\newblock {Learning rich features from RGB-D images for object detection and
  segmentation}.
\newblock In {\em ECCV}, 2014.

\bibitem{gupta2016cross}
Saurabh Gupta, Judy Hoffman, and Jitendra Malik.
\newblock {Cross Modal Distillation for Supervision Transfer}.
\newblock In {\em CVPR}, 2016.

\bibitem{kunwar2019u}
Saket Kunwar.
\newblock {U-Net Ensemble for Semantic and Height Estimation Using Coarse-Map
  Initialization}.
\newblock In {\em IGARSS}, 2019.

\bibitem{le20192019}
Bertrand Le~Saux, Naoto Yokoya, Ronny Hansch, Myron Brown, and Greg Hager.
\newblock {2019 Data Fusion Contest [technical committees]}.
\newblock {\em IEEE Geoscience and Remote Sensing Magazine}, 2019.

\bibitem{li2018megadepth}
Zhengqi Li and Noah Snavely.
\newblock {MegaDepth: Learning Single-View Depth Prediction from Internet
  Photos}.
\newblock In {\em CVPR}, 2018.

\bibitem{lin2017cascaded}
Di Lin, Guangyong Chen, Daniel Cohen-Or, Pheng-Ann Heng, and Hui Huang.
\newblock {Cascaded Feature Network for Semantic Segmentation of RGB-D Images}.
\newblock In {\em ICCV}, 2017.

\bibitem{liu2018see}
Shice Liu, Yu Hu, Yiming Zeng, Qiankun Tang, Beibei Jin, Yinhe Han, and Xiaowei
  Li.
\newblock {See and Think: Disentangling Semantic Scene Completion}.
\newblock In {\em NeurIPS}, 2018.

\bibitem{long2015fully}
Jonathan Long, Evan Shelhamer, and Trevor Darrell.
\newblock {Fully Convolutional Networks for Semantic Segmentation}.
\newblock In {\em CVPR}, 2015.

\bibitem{mou2018im2height}
Lichao Mou and Xiao~Xiang Zhu.
\newblock {IM2HEIGHT: Height Estimation from Single Monocular Imagery via Fully
  Residual Convolutional-Deconvolutional Network}.
\newblock {\em arXiv:1802.10249}, 2018.

\bibitem{park2017rdfnet}
Seong-Jin Park, Ki-Sang Hong, and Seungyong Lee.
\newblock {RDFNet: RGB-D Multi-level Residual Feature Fusion forIndoor Semantic
  Segmentation}.
\newblock In {\em ICCV}, 2017.

\bibitem{pintea2014deja}
Silvia~L Pintea, Jan~C van Gemert, and Arnold~WM Smeulders.
\newblock D{\'e}ja vu.
\newblock In {\em ECCV}, 2014.

\bibitem{qi20173d}
Xiaojuan Qi, Renjie Liao, Jiaya Jia, Sanja Fidler, and Raquel Urtasun.
\newblock {3D Graph Neural Networks for RGBD Semantic Segmentation}.
\newblock In {\em ICCV}, 2017.

\bibitem{schwarz2018rgb}
Max Schwarz, Anton Milan, Arul~Selvam Periyasamy, and Sven Behnke.
\newblock {RGB-D Object Detection and Semantic Segmentation for Autonomous
  Manipulation in Clutter}.
\newblock {\em IJRR}, 2018.

\bibitem{srivastava2017joint}
Shivangi Srivastava, Michele Volpi, and Devis Tuia.
\newblock Joint height estimation and semantic labeling of monocular aerial
  images with cnns.
\newblock In {\em IGARSS}, 2017.

\bibitem{vercauteren2009diffeomorphic}
Tom Vercauteren, Xavier Pennec, Aymeric Perchant, and Nicholas Ayache.
\newblock {Diffeomorphic demons: Efficient non-parametric image registration}.
\newblock {\em NeuroImage}, 2009.

\bibitem{walker2016uncertain}
Jacob Walker, Carl Doersch, Abhinav Gupta, and Martial Hebert.
\newblock {An Uncertain Future: Forecasting from Static Images using
  Variational Autoencoders}.
\newblock In {\em ECCV}, 2016.

\bibitem{walker2015dense}
Jacob Walker, Abhinav Gupta, and Martial Hebert.
\newblock {Dense Optical Flow Prediction from a Static Image}.
\newblock In {\em ICCV}, 2015.

\bibitem{wang2018depth}
Weiyue Wang and Ulrich Neumann.
\newblock {Depth-aware CNN for RGB-D Segmentation}.
\newblock In {\em ECCV}, 2018.

\bibitem{weir2019spacenet}
Nicholas Weir, David Lindenbaum, Alexei Bastidas, Adam~Van Etten, Sean
  McPherson, Jacob Shermeyer, Varun Kumar, and Hanlin Tang.
\newblock {SpaceNet MVOI: a Multi-View Overhead Imagery Dataset}.
\newblock In {\em ICCV}, 2019.

\bibitem{zampieri2018multimodal}
Armand Zampieri, Guillaume Charpiat, Nicolas Girard, and Yuliya Tarabalka.
\newblock Multimodal image alignment through a multiscale chain of neural
  networks with application to remote sensing.
\newblock In {\em ECCV}, 2018.

\bibitem{zheng2019pop}
Zhuo Zheng, Yanfei Zhong, and Junjue Wang.
\newblock {Pop-Net: Encoder-Dual Decoder for Semantic Segmentation and
  Single-View Height Estimation}.
\newblock In {\em IGARSS}, 2019.

\end{thebibliography}
}

\end{document}